\documentclass[10pt]{bathys}
\expandafter\ifx\csname iflatexml\endcsname\relax
  \expandafter\newif\csname iflatexml\endcsname
\fi
\usepackage[T1]{fontenc}

\usepackage{amsmath}
\usepackage{amssymb}
\usepackage{mathtools}
\usepackage{algorithm}
\usepackage{algorithmic}
\usepackage{booktabs}
\usepackage{multirow}
\usepackage{array}
\usepackage{graphicx}
\usepackage{float}
\usepackage{caption}
\usepackage{hyperref}
\usepackage{cleveref}

\newcommand{\agentname}{Bathys}
\newcommand{\modelnine}{Bathys 9B}
\newcommand{\modeltwentyseven}{Bathys 27B}
\newcommand{\sg}{\operatorname{sg}}

\title{Bathys: Advancing Deep Search Agents}

\iflatexml
\author{Zixuan Chen\textsuperscript{1} \and
  Shengyou Yuan\textsuperscript{1} \and
  Nang Yuan\textsuperscript{2} \and
  Ming Ma\textsuperscript{3} \and
  Zelin Xiao\textsuperscript{4} \and
  Lin Zhou\textsuperscript{5} \and
  Siyuan Luo\textsuperscript{6} \and
  Jiaqi Su\textsuperscript{7} \and
  Chengrui Wu\textsuperscript{8} \and
  Tiankuo Yao\textsuperscript{5} \and
  Nairong Zheng\textsuperscript{5} \and
  Cong Pang\textsuperscript{5,9} \and
  Jiapeng Luo\textsuperscript{5}}
\date{\textsuperscript{1}The University of Hong Kong \quad
  \textsuperscript{2}Shanghai Jiao Tong University \quad
  \textsuperscript{3}Harbin Institute of Technology \quad
  \textsuperscript{4}Peking University \quad
  \textsuperscript{5}SenseTime Research \quad
  \textsuperscript{6}University of Electronic Science and Technology of China \quad
  \textsuperscript{7}National University of Singapore \quad
  \textsuperscript{8}Zhejiang University \quad
  \textsuperscript{9}ShanghaiTech University \\
  November 2026 \quad Hugging Face: \url{https://huggingface.co/Bathys}}
\else
\author[1]{Zixuan Chen}
\author[1]{Shengyou Yuan}
\author[2]{Nang Yuan}
\author[3]{Ming Ma}
\author[4]{Zelin Xiao}
\author[5]{Lin Zhou}
\author[6]{Siyuan Luo}
\author[7]{Jiaqi Su}
\author[8]{Chengrui Wu}
\author[5]{Tiankuo Yao}
\author[5]{Nairong Zheng}
\author[5,9]{Cong Pang}
\author[5]{Jiapeng Luo}

\affiliation[1]{\afflogo{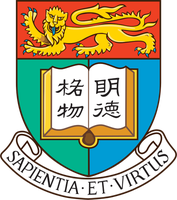}The University of Hong Kong}
\affiliation[2]{\afflogo{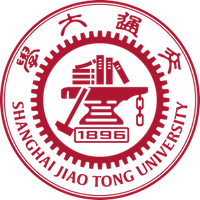}Shanghai Jiao Tong University}
\affiliation[3]{\afflogo{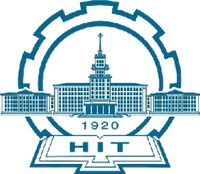}Harbin Institute of Technology}
\affiliation[4]{\afflogo{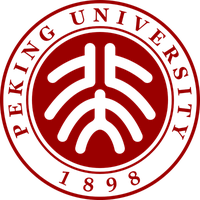}Peking University}
\affiliation[5]{\afflogo[0.72]{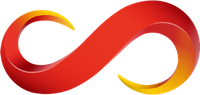}SenseTime Research}
\affiliation[6]{\afflogo{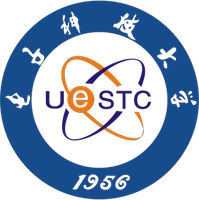}University of Electronic Science and Technology of China}
\affiliation[7]{\afflogo{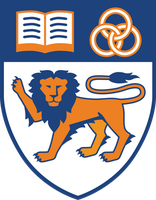}National University of Singapore}
\affiliation[8]{\afflogo{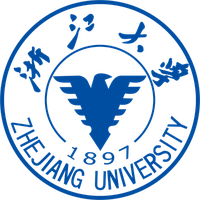}Zhejiang University}
\affiliation[9]{\afflogo{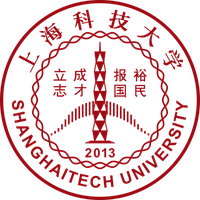}ShanghaiTech University}
\date{November 2026}
\checkdata[Hugging Face]{\url{https://huggingface.co/Bathys}}
\fi

\iflatexml\else
\abstract{
Deep search is a critical capability for agents tackling complex, open-ended tasks, where solving a problem requires iterative information gathering, query reformulation, and evidence-based verification rather than a single retrieval step. To address these requirements, we present \agentname{} 9B and 27B deep search agents built around a training framework that spans verifiable question synthesis, post-training, and online reinforcement learning (RL) under context compression. Our high-quality data construction ensures both search depth and reliability: each clue is reformulated as an indirect reference requiring independent retrieval, and retained evidence chains enable verification that the clues jointly determine a unique answer. Curated trajectories underpin a pipeline of supervised fine-tuning (SFT), Kahneman--Tversky Optimization (KTO), and Compaction-Managed Group Sequence Policy Optimization (CM-GSPO): SFT provides a stable foundation for tool use over long horizons and query reformulation; KTO uses binary feedback on trajectories to improve the selection of answers grounded in evidence; CM-GSPO optimizes execution sub-trajectories with length-normalized likelihood ratios on the recorded rollout contexts and reweights sub-trajectory advantages to correct the extra weighting introduced by independently averaging sub-trajectory losses. \agentname{} 9B and 27B attain 66.51\% and 80.49\% on BrowseComp, respectively, under an evaluation protocol with trajectory self-verification and conditional retries.
}
\fi

\begin{document}
\iflatexml
\begin{abstract}
Deep search is a critical capability for agents tackling complex, open-ended tasks, where solving a problem requires iterative information gathering, query reformulation, and evidence-based verification rather than a single retrieval step. To address these requirements, we present \agentname{} 9B and 27B deep search agents built around a training framework that spans verifiable question synthesis, post-training, and online reinforcement learning (RL) under context compression. Our high-quality data construction ensures both search depth and reliability: each clue is reformulated as an indirect reference requiring independent retrieval, and retained evidence chains enable verification that the clues jointly determine a unique answer. Curated trajectories underpin a pipeline of supervised fine-tuning (SFT), Kahneman--Tversky Optimization (KTO), and Compaction-Managed Group Sequence Policy Optimization (CM-GSPO): SFT provides a stable foundation for tool use over long horizons and query reformulation; KTO uses binary feedback on trajectories to improve the selection of answers grounded in evidence; CM-GSPO optimizes execution sub-trajectories with length-normalized likelihood ratios on the recorded rollout contexts and reweights sub-trajectory advantages to correct the extra weighting introduced by independently averaging sub-trajectory losses. \agentname{} 9B and 27B attain 66.51\% and 80.49\% on BrowseComp, respectively, under an evaluation protocol with trajectory self-verification and conditional retries.
\end{abstract}
\fi
\maketitle
\begin{figure}[H]
    \centering
    \includegraphics[width=\textwidth]{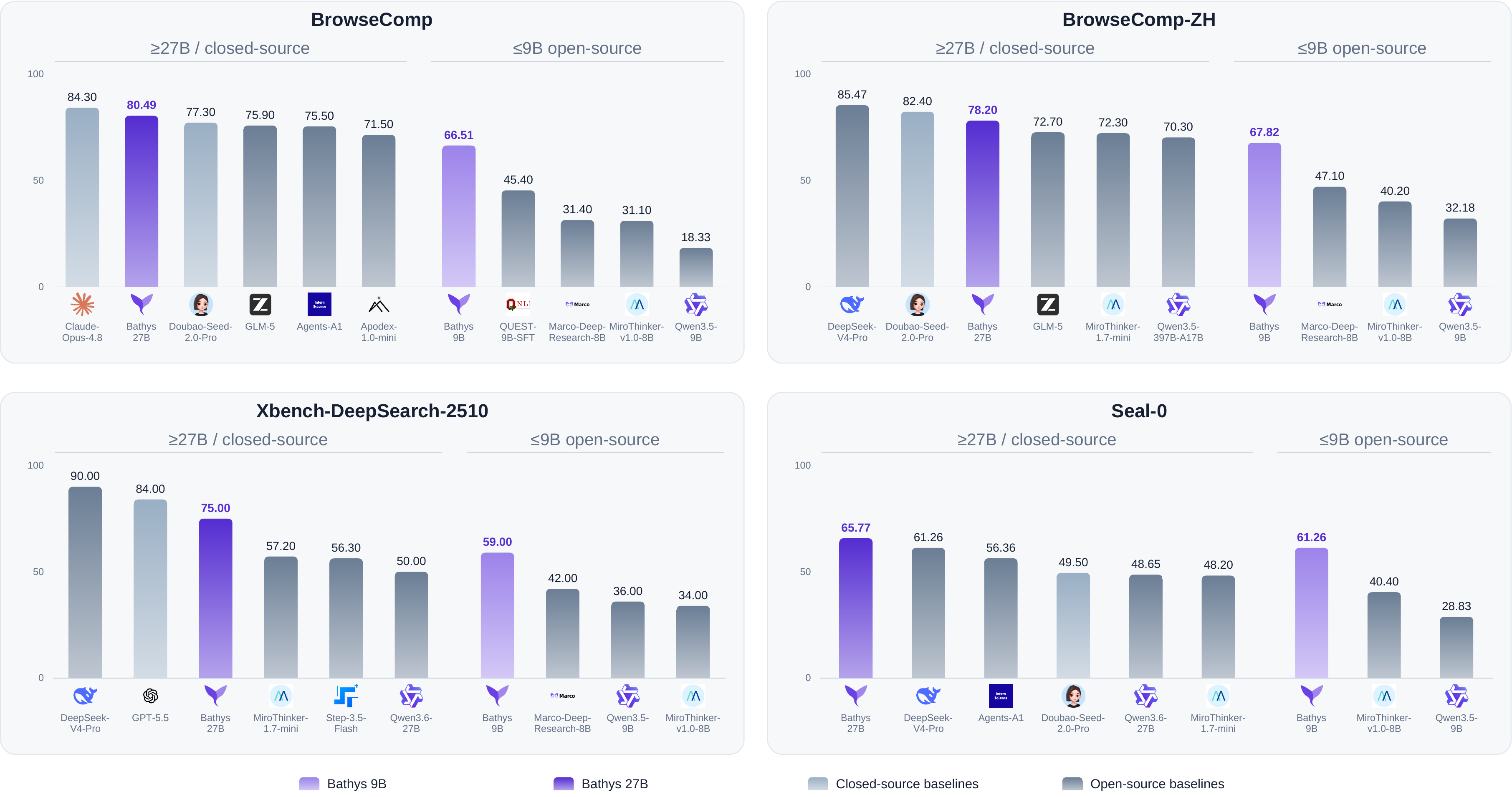}
    \caption{Main benchmark results for selected models from Table~\ref{tab:main_results}. Scores are accuracies (\%). Each panel shows a subset of the baselines, so Table~\ref{tab:main_results} includes additional entries, some of which score higher than \agentname{}. The \agentname{} models are highlighted in purple.}
    \label{fig:main_results_overview}
\end{figure}

\section{Introduction}
\label{sec:introduction}

Large language models are increasingly used as agents that plan, invoke tools, and revise intermediate conclusions over extended interactions. Deep search is a particularly demanding setting because retrieval and reasoning are tightly interleaved. Unlike a conventional search agent that retrieves information in a single pass, a deep search agent continuously refines its queries, synthesizes evidence across sources, and determines when the available evidence is sufficient to answer. The search trajectory is therefore part of the reasoning process itself, not merely a preprocessing step.

Recent work has made rapid progress on long-horizon deep search agents. Early methods enabled language models to invoke external tools and to interleave reasoning with actions~\cite{schick2023toolformer,yao2023react}. Reinforcement learning was then used to train search agents, first with local retrieval over static corpora~\cite{song2025r1searcher} and, more recently, with live web search~\cite{openai2025deepresearch,tongyidr,li2025webthinker,zheng2025deepresearcher}. Subsequent systems scale these agents along additional axes: the depth and frequency of agent--environment interaction~\cite{miromind2025mirothinker}, the interaction horizon and the breadth of domain expertise~\cite{bai2026agentsa1}, and the amount of verification performed at inference time~\cite{miromind2026mirothinker17,apodex2026}. In parallel, because existing question-answering datasets rarely require long-horizon exploration and challenging questions are costly to annotate manually, another line of work constructs harder training data and more controllable training environments. Representative approaches expand seed questions step by step under a set-theoretic formalization~\cite{tao2025webshaper}, obfuscate salient clues so that questions lack an obvious search entry point~\cite{liu2025webexplorer}, iteratively adjust question difficulty using execution feedback from a search agent~\cite{xu2026sagesteerableagenticdata}, and replace real search APIs with virtual tools~\cite{Chen2025ToolForgeAD} or a deterministic, verifiable search environment~\cite{geng2026deepsearch}. Recent systems have further combined verification-centric question and trajectory construction, test-time verification, staged post-training, reinforcement learning, and context management, with several approaches demonstrating strong performance at smaller model scales~\cite{zhu2026marco,xie2026quest,gao2025asearcher,wu2025resum,sun2025contextfolding,ye2025agentfold}. Nevertheless, benchmark performance can still vary substantially with the inference-time harness, rollout budget, context-management strategy, and training-data quality, making it difficult to determine whether gains in search depth are accompanied by equally strong data quality. We address this challenge with a deep-search system that spans data construction, post-training, online optimization, and evaluation.

We present \agentname{} as an integrated framework for constructing, training, and evaluating deep search agents, with the goal of improving both search capability and end-to-end verifiability. Our verifiable data pipeline constructs difficult questions with unique answers, explicit evidence chains, and controllable difficulty, providing auditable training data rather than search trajectories judged only by their final outputs. The subsequent SFT $\rightarrow$ KTO $\rightarrow$ CM-GSPO recipe establishes stable tool use, refines evidence-grounded decisions from binary trajectory feedback, and performs online optimization across context-compaction boundaries while correcting the extra weighting introduced by independently averaging sub-trajectory losses. We evaluate the resulting models with a shared Hermes-based harness, fixed web tools, an evaluation-only anti-cheating guard, and self-verification over the complete evidence chain. Building on the data-construction pipeline and CM-GSPO objective of Glasseek~\cite{pang2026glasseek}, this report integrates these components with SFT and KTO at 9B and 27B scales and evaluates the resulting agents. The data-construction method and CM-GSPO objective are adopted from that work. Under our evaluation protocol with self-verification, the 9B variant achieves accuracies of 66.51\%, 67.82\%, 59.00\%, and 61.26\% on BrowseComp, BrowseComp-ZH, Xbench, and Seal-0, respectively, demonstrating strong performance at a smaller scale. The 27B variant achieves 80.49\%, 78.20\%, 75.00\%, and 65.77\% on the same benchmarks, respectively.

\paragraph{Contributions and organization.}
This report presents the following system components and release plan:
\begin{itemize}
    \item \textbf{Verifiable data construction.} We describe seed selection,
joint-constraint construction, multi-hop evidence expansion, question fuzzification and repair,
and dual-agent verification. Following Glasseek~\cite{pang2026glasseek}, the construction makes a question multi-hop without
making it ill-posed. Each clue is rewritten as an indirect reference that has to
be resolved by a search of its own. Every rewrite retains an evidence chain back
to the source fact, so a verifier can confirm that the clues together still
single out one answer. Failed checks feed back into construction, which repairs
the offending constraint or retunes difficulty; added hops therefore cost the
solver more search rather than leaving the question ambiguous or unanswerable.

    \item \textbf{Context-aware post-training.} We detail trajectory curation and the SFT $\rightarrow$ KTO $\rightarrow$ CM-GSPO route: 128K-context SFT establishes tool-use behavior, and KTO incorporates binary feedback on evidence-supported search trajectories. For online optimization, we adopt CM-GSPO~\cite{pang2026glasseek} at both scales; it uses length-normalized sub-trajectory ratios to optimize the execution policy on recorded rollout contexts and applies sub-trajectory advantage reweighting to correct the extra weighting introduced by independently averaging sub-trajectory losses.

    \item \textbf{Open-source release.}
We release the \modelnine{} and \modeltwentyseven{} weights together with part of our post-training data: a subset of the curated SFT trajectories (Section~\ref{sec:trajectory_curation}) and a subset of the binary-labeled KTO preference data, in which whole trajectories are marked as desirable or undesirable (Section~\ref{sec:kto}). Both are available on Hugging Face at \url{https://huggingface.co/Bathys}.

\end{itemize}

\section{High-Quality Data Construction}
\label{sec:data}
\begin{figure}[H]
    \centering
    \includegraphics[width=\textwidth]{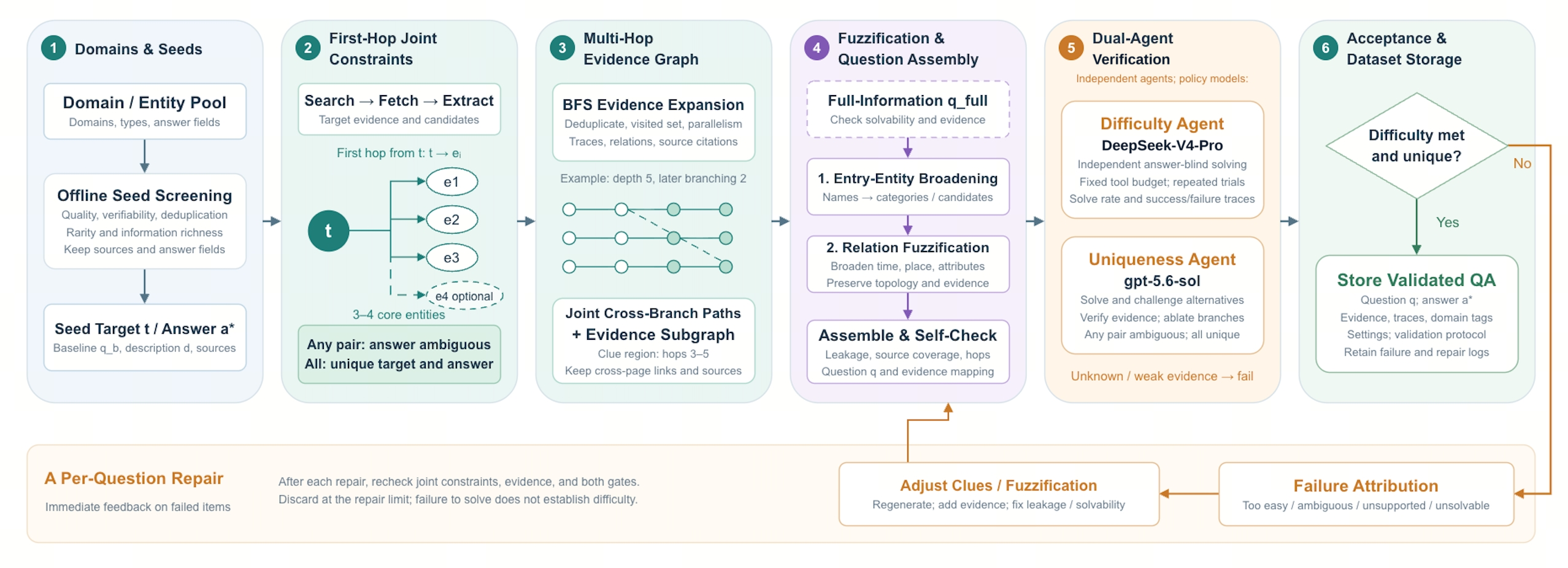}
    \caption{Overview of the data construction pipeline.
    Starting from screened seeds, we construct joint constraints,
    expand multi-hop evidence graphs, and apply fuzzification to
    assemble questions. Independent difficulty and uniqueness agents
    validate each question, with failed samples entering a bounded
    repair loop and accepted samples stored with supporting evidence
    and verification records.}
    \label{fig:data_pipeline}
\end{figure}

In this section, we introduce our high-quality data synthesis pipeline for constructing challenging, evidence-grounded deep-search questions, following the data construction of~\cite{pang2026glasseek}. The pipeline consists of seed selection, joint constraint construction, multi-hop evidence expansion, question fuzzification, and dual-agent verification, as illustrated in Figure~\ref{fig:data_pipeline}.

To begin with, consider the following illustrative question to introduce some basic elements of a question:

\begin{quote}
{In which year was the film released that featured an actor who played Ivan Locke in \textit{Locke}, an actor who portrayed Winston Churchill in \textit{Darkest Hour}, and an actor who played Charlie Croker in the 1969 film \textit{The Italian Job}?}
\end{quote}

The question contains three multi-hop clues, denoted by $\ell_1$, $\ell_2$, and $\ell_3$. Specifically, $\ell_1$ describes an actor who played Ivan Locke in \textit{Locke}, $\ell_2$ describes an actor who portrayed Winston Churchill in \textit{Darkest Hour}, and $\ell_3$ describes an actor who played Charlie Croker in the 1969 film \textit{The Italian Job}. These clues correspond to three entities: Tom Hardy ($e_1$), Gary Oldman ($e_2$), and Michael Caine ($e_3$), respectively. We refer to them as first-hop entities, as each actor is directly connected to the target film through an acting relationship. For each clue $\ell_i$, let $C_i$ denote the set of candidate films featuring its corresponding actor $e_i$. Thus, $C_1$, $C_2$, and $C_3$ represent the filmographies of Tom Hardy, Gary Oldman, and Michael Caine, respectively. While any two sets share multiple films with different release years, their intersection $\bigcap_{i=1}^{3} C_i=\{t\}$ identifies \textit{The Dark Knight Rises} as the target entity $t$ within the verified candidate scope. Since the question asks for the release year of this film, the ground-truth answer is $a^*=2012$.

In general, a synthesized question $q$ consists of $M$ multi-hop clues $\{\ell_i\}_{i=1}^{M}$, each describing a first-hop entity $e_i$ associated with a candidate target set $C_i$. Solving the question requires combining these candidate sets to identify the target entity $t$ and obtain its answer $a^*=f(t)$, where $f$ denotes the requested answer field. Our data synthesis pipeline follows the reverse process, as illustrated in Figure~\ref{fig:data_pipeline}. We first select a target entity $t$ and its ground-truth answer $a^*$, then identify $M\in\{3,4\}$ first-hop entities $\{e_i\}_{i=1}^{M}$ whose candidate sets $\{C_i\}_{i=1}^{M}$ satisfy the joint constraints. Starting from these entities, we progressively expand a multi-hop evidence graph $G=(V,E)$, construct a full-information reference question $q_{\mathrm{full}}$, and generate the final clues $\{\ell_i\}_{i=1}^{M}$ through fuzzification. These clues are assembled into the final question $q$, which subsequently undergoes dual-agent verification before acceptance and storage.

Accordingly, we first sample candidate targets from a pool spanning diverse domains, entity types, and answer fields. For each target $t$, we specify the answer field $f$, verify the normalized ground-truth answer $a^*=f(t)$, and collect an initial description with supporting sources. We then apply offline screening to remove duplicates, targets readily identifiable without retrieval, and seeds with unresolved identity ambiguity. Subsequent web search checks retrievability and evidence quality, retaining well-supported seeds for further construction.

For each retained seed, we then search for $M\in\{3,4\}$ distinct first-hop entities $\{e_i\}_{i=1}^{M}$ directly connected to $t$ through verified relations. Within the inspected candidate scope, $C_i$ contains targets satisfying the relation to $e_i$ and the shared question background. We select the first-hop entities such that
\begin{equation}
\begin{aligned}
|f(C_i \cap C_j)| &\geq 2,
    \qquad 1 \leq i < j \leq M,\\
\bigcap_{i=1}^{M} C_i &= \{t\},
\end{aligned}
\label{eq:joint_constraints}
\end{equation}
where $f(C)=\{f(x):x\in C\}$ denotes the set of distinct normalized answer values. Thus, each pair of branches retains at least one verified alternative whose answer differs from $a^*$, whereas all branches jointly identify $t$. If no verified combination meets these conditions, we revise the entity selection or discard the seed.

Once these constraints are established, we expand outward from each $e_i$ through budgeted breadth-first search, retrieving related entities, attributes, and relations through web search and page fetching. The collected information forms an evidence graph $G=(V,E)$, with supporting sources attached to the entities and relations. Deduplication, visited-node checks, and depth and branching limits control the expansion. We then jointly select complete evidence paths covering all core branches, preserving their connections to $t$ and any necessary shared-entity or cross-branch relations.

With these paths in place, we construct an internal full-information reference question $q_{\mathrm{full}}$, retaining the selected facts while masking target names, answer values, and revealing aliases. We subsequently apply two-stage fuzzification: first replacing specific outermost entity references with broader descriptions, and then relaxing identifying relations and attributes along the paths, such as replacing exact dates or locations with broader ranges. These transformations preserve factual support, relation direction, temporal consistency, and the entity to which each attribute belongs. The resulting clues $\{\ell_i\}_{i=1}^{M}$ are then assembled into the final question $q$ using only the approved descriptions.

Finally, two independent web-enabled agents evaluate the assembled question. The difficulty agent receives only $q$ and estimates its solve rate through repeated attempts under a fixed tool environment and budget; a question meets the difficulty requirement when this rate falls below a preset threshold. Independently, the uniqueness agent attempts a solution and searches for competing targets before inspecting the reference answer and supporting evidence for the audit. We also recheck the joint constraints using only the information visible in $q$: all two-branch variants are examined for evidence-backed alternative answers and unintended shortcuts, while shared-entity and cross-branch relations are checked jointly. Acceptance requires the prescribed difficulty level, a verified evidence-supported solution, factual grounding, and a completed audit finding no valid competing target within the inspected candidate scope. Unresolved candidates or incomplete evidence do not count as passes.

When a question fails validation, solver traces and counterexamples guide a bounded repair loop. Overly revealing clues are weakened, missing discriminative facts are restored, and unsupported relations are corrected or supplemented with evidence. The question is then regenerated and rechecked against all acceptance criteria; samples that still fail at the repair limit are discarded. Accepted pairs $(q,a^*)$ are stored together with their supporting evidence, candidate checks, verification traces, repair histories, and protocol settings.

\FloatBarrier
\section{Post-Training: SFT, KTO, and CM-GSPO}
\label{sec:training}

Our post-training route is organized as SFT $\rightarrow$ KTO $\rightarrow$ CM-GSPO. SFT teaches a stable interaction format and broad search behavior. KTO then serves as the selected offline preference stage, improving evidence integration, candidate discrimination, and answer calibration from fixed binary-feedback trajectories. Direct Preference Optimization (DPO) is trained on the same source pool as a paired-preference control; KTO yields the stronger checkpoint on these dimensions and is the stage carried into CM-GSPO. CM-GSPO performs online optimization using fresh rollouts and explicitly accounts for the context states created during execution.

\subsection{Trajectory Curation}
\label{sec:trajectory_curation}

For each training question, one or more teacher policies generate complete interaction trajectories containing assistant reasoning, tool calls, tool observations, and the final answer. The raw trajectories pass through a deterministic quality funnel that validates the message and tool-call schema, filters repetitive or no-progress searches and abnormal tool behavior, verifies answer-bearing claims against collected evidence, and checks length and packing boundaries. Only structurally valid, evidence-grounded, and trainable trajectories are retained.

\subsection{Supervised Fine-Tuning}

We apply full-parameter SFT as the first post-training stage. SFT establishes stable tool use, query decomposition and reformulation, evidence integration, error recovery, and concise answer generation over long search trajectories.

\paragraph{Data mixture.}
Both scales follow the same data methodology and essentially the same data recipe. Differences in sample counts, sampling weights, or optimization budget are implementation details and do not change the data methodology. The shared retention data preserves broad capabilities in tool use, multi-step reasoning, planning, function calling, and mathematics.

The SFT loss is the standard masked causal objective over assistant policy tokens:
\begin{equation}
    \mathcal L_{\mathrm{SFT}}(\theta)
    =-\frac{1}{\sum_n |M_n|}\sum_n\sum_{t\in M_n}
    \log\pi_\theta(y_{n,t}\mid x_n,y_{n,<t}),
\end{equation}
where $M_n$ excludes user content, tool observations, padding, and any context tokens that should not receive policy gradients. The 9B and 27B runs share the same data schema, loss definition, and data-quality criteria; scale-specific implementation settings are reported separately in Appendix~\ref{app:training_details}.

\paragraph{Implementation Details.}
We initialize the 9B and 27B agents from the open-source Qwen3.5-9B and Qwen3.6-27B checkpoints, respectively, and train both models with full-parameter SFT using ms-swift and Megatron-LM. Both runs use a maximum sequence length of 128K tokens, a micro-batch size of 1, a global batch size of 64, and a 3\% warmup ratio. Over-length examples are removed before training, while retained long sequences use packing, padding-free batching, full activation recomputation, and Flash Attention. Parallelism and other SFT launch settings are reported in Table~\ref{tab:sft_training_details} in Appendix~\ref{app:training_details}.

\subsection{KTO}
\label{sec:kto}

\subsubsection{Algorithm Overview}

Our post-training stack first applies offline preference optimization and then online reinforcement learning. After SFT, we compare KTO~\cite{ethayarajh2024kto} with a DPO baseline built from the same preference source pool. In our experiments, KTO provides the stronger and more stable preference checkpoint, so the main route uses SFT $\rightarrow$ KTO before entering CM-GSPO.

\subsubsection{KTO as the Main Offline Preference Stage}

The preference source pool combines two complementary signals: RL feedback rollouts from our search-agent iterations, and in-house contrast data where stronger-model trajectories are contrasted with negative or hard-negative search trajectories. KTO consumes this pool as binary desirability feedback, while DPO views the same pool through chosen/rejected pairs.

Desirable examples are complete, evidence-supported, tool-valid, and consistent between the final answer and the retrieved evidence. Undesirable examples include wrong answers, empty answers, timeouts, malformed tool use, unsupported conclusions, premature anchoring on a locally plausible but globally invalid candidate, and unnecessary over-search after sufficient evidence has already been collected. This labeling scheme makes the preference stage directly target the behavior we want before online RL: integrate distributed constraints, revise candidates when new evidence contradicts the current hypothesis, and stop with a calibrated final answer.

Let $a\in\{+1,-1\}$ denote a desirable or undesirable label and let
\begin{equation}
    u_\theta(x,y)=\beta\log\frac{\pi_\theta(y\mid x)}{\pi_{\mathrm{ref}}(y\mid x)}-z_{\mathrm{ref}},
\end{equation}
where the Kullback--Leibler (KL)-divergence-derived reference offset is
\begin{equation}
    z_{\mathrm{ref}}
    =\mathbb E_{x'\sim\mathcal D}\!\left[
    \beta D_{\mathrm{KL}}\!\left(\pi_\theta(\cdot\mid x')\,\|\,\pi_{\mathrm{ref}}(\cdot\mid x')\right)
    \right].
\end{equation}
A compact form of the binary-feedback objective is
\begin{equation}
    \mathcal L_{\mathrm{KTO}}(\theta)
    =\mathbb E_{(x,y,a)}\!\left[
    \lambda_a\left(1-\sigma\!\left(a\,u_\theta(x,y)\right)\right)
    \right],
\end{equation}
with separate weights for desirable and undesirable examples.

\paragraph{Implementation Details.}
KTO is initialized from the corresponding scale-matched SFT checkpoint. The training uses a maximum sequence length of 128K tokens, two training epochs, a learning rate of $5\times10^{-7}$ with a 5\% warmup ratio, and micro/global batch sizes of 1/16. We set $\beta=0.1$ and weight desirable and undesirable trajectories equally. Training uses bfloat16 (BF16), sequence packing, full activation recomputation, and Flash Attention, and removes examples that exceed the length limit. Table~\ref{tab:preference_training_details} summarizes the preference-optimization configuration.

\paragraph{DPO selection baseline.}
For comparison, we convert the same source pool into chosen/rejected trajectory pairs by pairing complete, evidence-supported trajectories with representative search or answer failures. The two objectives use the same preference source pool and matched evaluation settings, but consume different training units: unpaired labeled trajectories for KTO and chosen/rejected pairs for DPO. The comparison therefore evaluates the resulting preference-training recipes rather than isolating the objective independently of data representation. KTO yields the better preference checkpoint on our target metrics and is therefore used to initialize CM-GSPO.

\subsection{CM-GSPO}
\label{sec:cm_gspo}
 
During online training, a search rollout may exhaust the context window before producing a final answer. Compression allows the search to continue, but changes the history available to subsequent actions. The policy update must therefore account for the contexts used during generation and for the way the rollout is split into training sequences. Compaction-Managed GSPO (CM-GSPO)~\cite{pang2026glasseek} extends Group Sequence Policy Optimization~\cite{zheng2025gspo} to meet both requirements: it computes importance ratios within compression-defined sub-trajectories and normalizes their contributions to retain the weighting of the original rollout.
 
\paragraph{Representing compressed rollouts.}
For a query $x$, the context manager $\mathcal C$ reduces the running history whenever its length exceeds a threshold $T$. This can be implemented through observation masking, summarization, or an adaptive compression policy~\cite{kang2025acon,verma2026activecontextcompression}. We record each compression event as a boundary in rollout $y_i$, yielding $K_i$ consecutive segments:
\begin{equation}
    y_i=y_{i,1}\circ y_{i,2}\circ\cdots\circ y_{i,K_i}.
\end{equation}
Each segment carries a policy-token mask $\mathcal M_{i,k}$. Its optimization length is $L_{i,k}=|\mathcal M_{i,k}|$, and the rollout contains $L_i=\sum_{k=1}^{K_i}L_{i,k}$ such tokens in total. Tool observations and compressed-context tokens supply conditioning information but are excluded from this mask.
 
The state passed from segment $k$ to the next is denoted by $h_{i,k}$; $h_{i,0}$ contains the initial query. With $o_{i,k}$ representing the tool observations collected in that segment, the context transition is
\begin{equation}
    h_{i,k}\sim q_{\mathcal C}(\cdot\mid h_{i,k-1},y_{i,k},o_{i,k}),
\end{equation}
Here, $q_{\mathcal C}$ describes how the next conditioning state is constructed. The policy objective below optimizes the generated search actions, so this context-transition distribution is excluded from its importance ratios.
 
\paragraph{Importance correction under recorded contexts.}
When updating the policy, we evaluate each segment using the context saved at its starting boundary. Write $p^\theta_{i,k,t}=\pi_\theta(y_{i,k,t}\mid x,h_{i,k-1},y_{i,k,<t})$ for the current-policy probability and $p^{\theta_{\mathrm{old}}}_{i,k,t}$ for the behavior-policy probability under the same conditioning history. Following the sequence-ratio construction of GSPO, we take the geometric mean of the probability ratios over the segment's unmasked tokens:
\begin{equation}
    s^{\mathrm{sub}}_{i,k}(\theta)
    =\exp\!\left[
    \frac{1}{L_{i,k}}\sum_{t\in\mathcal M_{i,k}}
    \log\frac{p^\theta_{i,k,t}}{p^{\theta_{\mathrm{old}}}_{i,k,t}}
    \right],
    \label{eq:sub_ratio}
\end{equation}
To use this segment-level correction in a token-wise loss, we introduce a surrogate for each $t\in\mathcal M_{i,k}$:
\begin{equation}
    \widetilde s^{\mathrm{sub}}_{i,k,t}(\theta)
    =\sg\!\left[s^{\mathrm{sub}}_{i,k}(\theta)\right]
    \exp\!\left(\log p^\theta_{i,k,t}-\sg[\log p^\theta_{i,k,t}]\right).
\end{equation}
The operator $\sg[\cdot]$ stops gradient propagation. Thus, $\widetilde s^{\mathrm{sub}}_{i,k,t}$ has the same forward value as $s^{\mathrm{sub}}_{i,k}$, while its gradient flows through the current token's log probability. This preserves the shared segment ratio when applying the clipped policy update.
 
\paragraph{Sub-trajectory advantage reweighting.}
Segmentation changes the unit used for importance correction, but the training signal still comes from the outcome of the complete search. All segments of rollout $i$ inherit its terminal reward $r_i$ and group-normalized advantage $\widehat A_i$. Giving every segment its own unweighted token average would amplify short segments and increase the contribution of rollouts split more often. We counter this effect by scaling the shared advantage according to each segment's fraction of the rollout's policy tokens:
\begin{equation}
    \widehat A^{\mathrm{cm}}_{i,k}=\frac{L_{i,k}}{L_i}\widehat A_i.
    \label{eq:cm_advantage}
\end{equation}
Since $L_{i,k}/L_i$ is positive, it can be absorbed into the advantage inside the clipped objective. For a group of $G$ rollouts and clipping parameter $\epsilon$, the resulting update maximizes
\begin{equation}
\begin{aligned}
J_{\mathrm{CM\text{-}GSPO}}(\theta)=\mathbb E\bigg[\frac{1}{G}\sum_{i=1}^{G}\sum_{k=1}^{K_i}
\frac{1}{L_{i,k}}\sum_{t\in\mathcal M_{i,k}}
\min\Big(&\widetilde s^{\mathrm{sub}}_{i,k,t}(\theta)\widehat A^{\mathrm{cm}}_{i,k},\\
&\operatorname{clip}(\widetilde s^{\mathrm{sub}}_{i,k,t}(\theta),1-\epsilon,1+\epsilon)
\widehat A^{\mathrm{cm}}_{i,k}\Big)\bigg].
\end{aligned}
\label{eq:cm_objective}
\end{equation}
The identity $(1/L_{i,k})(L_{i,k}/L_i)=1/L_i$ gives every optimized token within rollout $i$ the same outer normalization. This removes the extra weighting introduced by independently averaging its sub-trajectories, while the sub-trajectory ratios and clipping retain their dependence on the chosen segmentation.
 
\paragraph{Training implementation.}
For each model scale, online training starts from its KTO checkpoint and collects fresh interactions in the agent environment. The rollout collector saves compression boundaries and their associated context states so that each segment can be reconstructed as a training sequence. The optimizer applies the policy-token mask and uses the original rollout length and shared advantage when combining segment losses. The implementation uses separate lower and upper clipping values and an additional token-level importance correction, as listed in Table~\ref{tab:cm_gspo_training_details}.

\FloatBarrier
\section{Results and Analysis}
\label{sec:results}

\subsection{Evaluation Setup}
\label{subsec:evaluation-setup}
\paragraph{Benchmarks.}
Our main evaluation covers four benchmarks spanning English and Chinese web search, end-to-end deep search, and reasoning over conflicting evidence. BrowseComp~\citep{wei2025browsecomp} measures agents' ability to persistently navigate the internet in search of hard-to-find, entangled information, while BrowseComp-ZH (BC-ZH)~\citep{browsecompzh2025} evaluates large language model (LLM) agents' web-browsing and reasoning capabilities in the Chinese information ecosystem. Xbench-DeepSearch-2510 (abbreviated as Xbench)~\citep{chen2025xbench} assesses end-to-end deep-search ability on Chinese-language questions, requiring agents to integrate planning, search, reasoning, and summarization. Seal-0~\citep{pham2025sealqa} evaluates search-augmented language models on fact-seeking questions where web search produces conflicting, noisy, or unhelpful results, with an emphasis on factual accuracy and reasoning.

\paragraph{Implementation Details.}
To evaluate the generalization and adaptability of the model to agentic architectures of general use, we adopt Hermes\footnote{\url{https://github.com/nousresearch/hermes-agent}} as our evaluation harness. We equip the agent with web tools to complement its native capabilities, including code execution, vision, and sub-agent delegation. Specifically, we implement a search tool powered by Serper\footnote{\url{https://serper.dev/}}, which returns the top 10 Google search results, and a fetch tool built on Jina\footnote{\url{https://jina.ai/reader/}} for webpage reading. This unified, multi-tool configuration tests the model's intrinsic reasoning and tool-use capabilities under a standardized and versatile framework.

To mitigate benchmark leakage, we enable an evaluation-only anti-cheating guard that blocks access to all Hugging Face pages, including \texttt{huggingface.co}, \texttt{hf.co}, and associated content-delivery domains. The guard also rejects searches, URLs, and retrieved content containing benchmark-specific identifiers, regardless of the hosting website, and filters results that substantially reproduce the evaluation question or expose reference answers. This policy is applied across search, extraction, browser, and vision tools.

\paragraph{Evaluation Protocol.}
We evaluate both \agentname{} models with the same agent harness and tool configuration across the four benchmarks. Baseline scores are taken from official reports and released evaluation results, except for the self-measured entries marked in Table~\ref{tab:main_results}. For the \agentname{} main results, inspired by the global verification approach of the MiroThinker-1.7 \& H1 report~\cite{miromind2026mirothinker17}, we run the main workflow once, then ask the same model to perform self-verification and assess the resulting trajectory for logical consistency and evidential support. A retry is triggered only if the trajectory fails this check; otherwise, its answer is directly used as the final prediction. If retries are required, the answer from the first trajectory that passes self-verification is used for benchmark scoring.

The main results measure the complete model--harness system with self-verification. The training-stage comparisons in Tables~\ref{tab:kto_dpo_scale_results} and~\ref{tab:ablation_gspo} omit self-verification and are analyzed separately. Appendix~\ref{app:browsecomp_ambiguity} documents reference-answer variants and a potentially ambiguous BrowseComp item.

\paragraph{BrowseComp analysis subset.}
For trajectory analysis and ablation experiments, we randomly sample 200 BrowseComp questions and keep this subset fixed across the compared checkpoints, distinct from the full 1,266-question evaluation in Table~\ref{tab:main_results}. Figure~\ref{fig:browsecomp-trajectory-reliability} and Tables~\ref{tab:kto_dpo_scale_results} and~\ref{tab:ablation_gspo} report results on this subset without self-verification. Subset scores are not directly comparable to full-set BrowseComp scores.

\subsection{Main Results}

\paragraph{Performance.}
Table~\ref{tab:main_results} summarizes the main results; Figure~\ref{fig:main_results_overview} visualizes selected entries. \modelnine{} achieves 66.51 on BrowseComp, 67.82 on BrowseComp-ZH, 59.00 on Xbench, and 61.26 on Seal-0, improving over its Qwen3.5-9B base by 48.18, 35.64, 23.00, and 32.43 percentage points, respectively, and exceeding the scores of all open-source baselines at or below 9B listed in the table on each benchmark with available results. \modeltwentyseven{} reaches 80.49, 78.20, 75.00, and 65.77 on the same benchmarks, exceeding the Qwen3.6-27B baseline by 28.52, 25.95, 25.00, and 17.12 points. On BrowseComp, it is surpassed by two open-source entries in the table, DeepSeek-V4-Pro (83.40) and Kimi-K2.6 (83.20), which lead by less than 3 points. Baseline scores not marked as self-measured are reproduced from official reports and released evaluations under their respective protocols.

\agentname{} shows its strongest relative result on Seal-0, whose questions are ones for which web search returns conflicting, noisy, or unhelpful results. \modelnine{} reaches 61.26, matching DeepSeek-V4-Pro and above every other baseline, including Agents-A1 (56.36), with a 20.86-point margin over the next open-source model at or below 9B. \modeltwentyseven{} scores 65.77, the highest in Table~\ref{tab:main_results}, 4.51 points above DeepSeek-V4-Pro. In the separate training-stage comparison in Table~\ref{tab:ablation_gspo}, CM-GSPO raises Seal-0 from 49.55 after KTO to 62.16 without self-verification.

\paragraph{FutureX leaderboard.}
We evaluate \agentname{} on the FutureX weekly leaderboard, using the official Week 3 snapshot from August 2026. FutureX is a dynamic, live benchmark for future prediction: its tasks require agents to gather current information, synthesize evidence from multiple sources, reason under uncertainty, and forecast outcomes across domains such as politics, economics, and finance~\cite{futurex2025}. Stronger search capability can therefore provide a more timely and complete information basis for prediction; our result is consistent with this positive relationship, although the leaderboard does not isolate search quality from reasoning, calibration, or other agent capabilities.

\begin{table}[t]
\centering
\small
\caption{Main benchmark results. Scores are accuracies (\%). \textsuperscript{\S} marks scores we measured ourselves under the harness and tool stack described above, because the corresponding releases do not report these benchmarks; all other baseline scores are reproduced from official model reports and released evaluation results. Unavailable baseline model sizes and benchmark scores are shown as --. The \agentname{} rows are bolded and use the self-verification protocol described above; Tables~\ref{tab:kto_dpo_scale_results} and~\ref{tab:ablation_gspo} report scores without self-verification. Our full evaluation sets contain 1,266 (BrowseComp), 289 (BC-ZH), 100 (Xbench-DeepSearch-2510), and 111 (Seal-0) questions.}
\label{tab:main_results}
\scalebox{0.95}{%
\begin{tabular}{lccccc}
\toprule
Model & Size & BrowseComp & BC-ZH & Xbench-DeepSearch-2510 & Seal-0 \\
\midrule
\multicolumn{6}{l}{\textbf{Closed-Source Models}} \\
\midrule
GPT-5.5 & -- & 84.40 & -- & 84.00 & -- \\
Claude-Sonnet-5 & -- & \textbf{84.70} & -- & -- & -- \\
Claude-Opus-4.8 & -- & 84.30 & -- & -- & -- \\
Doubao-Seed-2.0-Pro & -- & 77.30 & 82.40 & -- & 49.50 \\
\midrule
\multicolumn{6}{l}{\textbf{Open-Source Models ($\geq$27B)}} \\
\midrule
DeepSeek-V4-Pro & 1.6T & 83.40 & \textbf{85.47}\textsuperscript{\S} & \textbf{90.00} & \textbf{61.26}\textsuperscript{\S} \\
Kimi-K2.6 & 1T & 83.20 & -- & -- & -- \\
GLM-5 & 744B & 75.90 & 72.70 & -- & -- \\
Qwen3.5-397B-A17B & 397B & 78.60 & 70.30 & -- & 46.90 \\
Step-3.5-Flash & 199B & 69.00 & 66.90 & 56.30 & -- \\
MiroThinker-1.7-mini & 30B & 67.90 & 72.30 & 57.20 & 48.20 \\
Agents-A1 & 35B & 75.50 & -- & 86.00 & 56.36 \\
Apodex-1.0-mini & 35B & 71.50 & 80.60 & -- & -- \\
Qwen3.6-27B & 27B & 51.97\textsuperscript{\S} & 52.25\textsuperscript{\S} & 50.00\textsuperscript{\S} & 48.65\textsuperscript{\S} \\
\midrule
\multicolumn{6}{l}{\textbf{Open-Source Models ($\leq$9B)}} \\
\midrule
Qwen3.5-9B & 9B & 18.33\textsuperscript{\S} & 32.18\textsuperscript{\S} & 36.00\textsuperscript{\S} & 28.83\textsuperscript{\S} \\
MiroThinker-v1.0-8B & 8B & 31.10 & 40.20 & 34.00 & 40.40 \\
Marco-DeepResearch-8B & 8B & 31.40 & 47.10 & 42.00 & -- \\
QUEST-9B-SFT & 9B & 45.40 & -- & -- & -- \\
\midrule
\multicolumn{6}{l}{\textbf{\agentname{} Models}} \\
\midrule
\textbf{\modelnine} & \textbf{9B} & \textbf{66.51} & \textbf{67.82} & \textbf{59.00} & \textbf{61.26} \\
\textbf{\modeltwentyseven} & \textbf{27B} & \textbf{80.49} & \textbf{78.20} & \textbf{75.00} & \textbf{65.77} \\
\bottomrule
\end{tabular}
}
\end{table}

\begin{table}[t]
\centering
\small
\caption{FutureX Weekly Leaderboard results (Week 3, August 2026). Scores and harness names are taken from the official leaderboard snapshot; submission identifier Maxmilia\_v3 corresponds to \modeltwentyseven{}. Harness denotes the FutureX ``Agent Framework'' field.}
\label{tab:futurex}
\scalebox{0.95}{%
\begin{tabular}{llcccccc}
\toprule
Model & Harness & Rank & L1 & L2 & L3 & L4 & Score \\
\midrule
\modeltwentyseven{} (Maxmilia\_v3) & hermes & \textbf{1} & 100.00 & 60.00 & 70.61 & 72.20 & \textbf{72.06} \\
GLM-5.2 & glm52-tools & 2 & 100.00 & 60.00 & 75.15 & 59.86 & 68.49 \\
DeepSeek-V4-Pro & JSearch\_V1\_S4 & 3 & 100.00 & 40.00 & 84.24 & 59.77 & 67.18 \\
\bottomrule
\end{tabular}
}
\end{table}

As shown in Table~\ref{tab:futurex}, \modeltwentyseven{} ranks first on the FutureX Week 3 leaderboard from August 2026, achieving an overall score of 72.06 and outperforming GLM-5.2 (68.49) and DeepSeek-V4-Pro (67.18). The score is computed over 46 events spanning four difficulty levels. Following the FutureX aggregation rule~\cite{futurex2025}, the overall score is $0.1L_1+0.2L_2+0.3L_3+0.4L_4$, where $L_j$ denotes the score at difficulty level $j$.

\paragraph{Capability-level analysis.}
To relate these results to the recipe, we also compare the trajectories of the checkpoints of every training stage on BrowseComp questions. All checkpoints use identical harness, tool, context-management, and budget settings. We measure completion, timeout, and context-limit failure rates, together with trajectory length summarized by the mean, median, and 90th-percentile number of rounds.

\begin{figure}[H]
    \centering
    \includegraphics[width=0.80\linewidth]{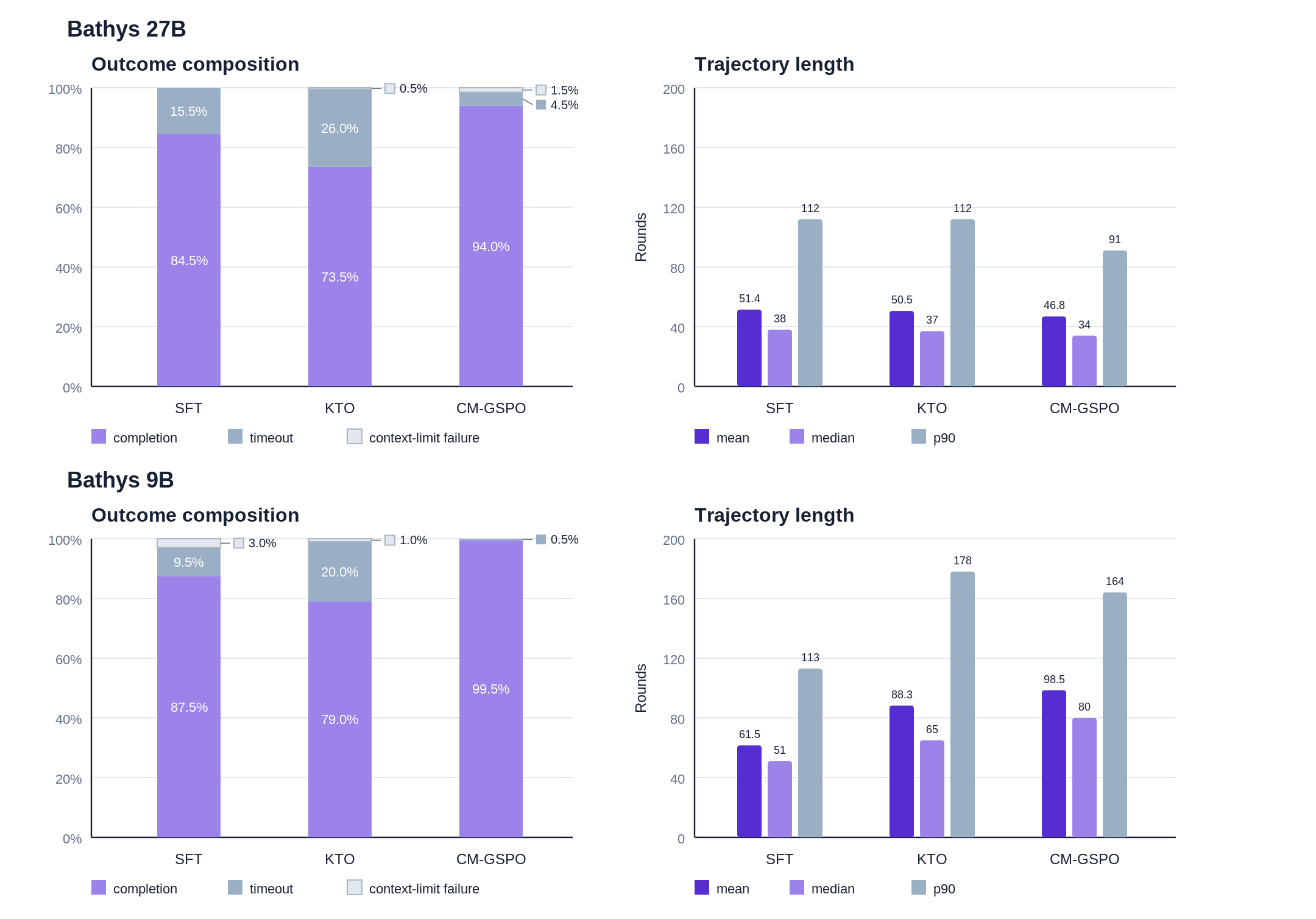}
    \caption{Trajectory reliability and cost across training stages. Rows separate \modeltwentyseven{} and 9B checkpoints; columns show outcome composition and trajectory length. Results are measured on the fixed 200-question subset of BrowseComp, without the self-verification approach.}
    \label{fig:browsecomp-trajectory-reliability}
\end{figure}

As summarized in Figure~\ref{fig:browsecomp-trajectory-reliability}, the training stages change reliability and trajectory cost in different ways at the two model scales. Although KTO introduces a self-checking preference and is associated in the trajectory figure with lower completion and higher timeout than SFT at both scales, its judged BrowseComp accuracy improves over SFT at both scales. At 27B, completion falls from 84.5\% to 73.5\% and timeout rises from 15.5\% to 26.0\%, while mean and median rounds remain close to SFT and the 90th percentile (p90) remains 112. Nevertheless, judged accuracy rises from 65.00\% for SFT to 66.50\% after SFT $\rightarrow$ KTO (+1.50 percentage points). CM-GSPO raises completion to 94.0\% and reduces timeout to 4.5\%. It also shortens the trajectory distribution relative to SFT, reducing mean rounds from 51.4 to 46.8 and p90 rounds from 112 to 91. At 9B, KTO completion falls from 87.5\% to 79.0\% and timeout rises from 9.5\% to 20.0\%; its long tail grows substantially, with p90 rounds increasing from 113 to 178, while judged accuracy increases from 53.50\% for SFT to 57.00\% after SFT $\rightarrow$ KTO (+3.50 points). CM-GSPO reaches 99.5\% completion and 0.5\% timeout, but its trajectories are longer than SFT, with mean rounds increasing from 61.5 to 98.5 and p90 rounds from 113 to 164. Overall, the figure and the judged accuracies support a clear reliability improvement for the final stage at both scales, together with a cost reduction at 27B and a longer but more reliable trajectory distribution at 9B.

\subsection{KTO Preference Optimization Analysis}

This subsection compares the offline preference stage at 9B and 27B scales. Table~\ref{tab:kto_dpo_scale_results} reports BrowseComp accuracy for scale-matched SFT, SFT $\rightarrow$ KTO, and SFT $\rightarrow$ DPO checkpoints. The $\Delta$ values denote the change in accuracy relative to the corresponding SFT checkpoint, measured in percentage points. The DPO rows provide a matched baseline for comparing preference-stage choices.

\begin{table}[H]
\centering
\small
\caption{BrowseComp accuracy (\%) for offline preference optimization, measured on a fixed 200-question subset of BrowseComp. Results are obtained without the self-verification approach. $\Delta$ is relative to SFT, in percentage points.}
\label{tab:kto_dpo_scale_results}
\scalebox{0.95}{%
\begin{tabular}{llp{0.30\textwidth}cc}
\toprule
Scale & Role & Checkpoint / comparison & \begin{tabular}[b]{@{}c@{}}Accuracy (\%)\end{tabular} & $\Delta$ vs. SFT \\
\midrule
9B & Main route & SFT & 53.50 & -- \\
9B & Selection baseline & SFT $\rightarrow$ DPO & 49.50 & $-4.00$ \\
9B & Main route & SFT $\rightarrow$ KTO & 57.00 & +3.50 \\
\midrule
27B & Main route & SFT & 65.00 & -- \\
27B & Selection baseline & SFT $\rightarrow$ DPO & 63.50 & $-1.50$ \\
27B & Main route & SFT $\rightarrow$ KTO & 66.50 & +1.50 \\
\bottomrule
\end{tabular}
}
\end{table}

The comparison provides a consistent selection signal across model scales. At 9B, KTO raises BrowseComp accuracy from 53.50\% to 57.00\% (+3.50 percentage points), whereas DPO lowers it to 49.50\% ($-4.00$ points). The resulting 7.50-point gap between KTO and DPO corresponds to 15 additional correct answers on this 200-question subset. The ordering is preserved at 27B: KTO remains positive (+1.50 points), while DPO remains negative ($-1.50$ points), yielding a 3.00-point KTO--DPO gap. KTO is the only preference stage in this comparison that improves its scale-matched SFT checkpoint at both model sizes. The 27B improvement over SFT corresponds to three additional correct answers on the subset; these point estimates do not establish statistical significance.

\FloatBarrier
 
\subsection{CM-GSPO Ablation}
 
\begin{table}[H]
\centering
\small
\caption{Ablation study of online training strategies for \modeltwentyseven{}, with KTO as the baseline. Scores are accuracies (\%), obtained without the self-verification approach. BrowseComp scores are measured on a fixed 200-question subset of BrowseComp.}
\label{tab:ablation_gspo}
\resizebox{\linewidth}{!}{%
\begin{tabular}{lcccc}
\toprule
Method & BrowseComp & BrowseComp-ZH & Seal-0 & Xbench-DeepSearch-2510 \\
\midrule
KTO (without the self-verification approach)  & 66.50 & 75.78 & 49.55 & 62.00 \\
Vanilla GSPO (without the self-verification approach) & 66.00 & 65.74 & 43.24 & 52.00 \\
Compaction RL (without the self-verification approach) & 64.00 & 59.86 & 42.34 & 51.00\\
\rowcolor{gray!15}
CM-GSPO (without the self-verification approach) & 75.50 & 71.97 & 62.16 & 70.00 \\
\bottomrule
\end{tabular}
}
\end{table}
\FloatBarrier
 
Vanilla GSPO is trained without context compaction, and Compaction RL incorporates context compaction into GSPO training. CM-GSPO replaces GSPO with the objective of Section~\ref{sec:cm_gspo} while retaining context compaction.
 
Neither GSPO variant improves on KTO. Vanilla GSPO is close to KTO on BrowseComp (0.50 points lower) but trails it by 6.31 to 10.04 points on the other three benchmarks. Compaction RL trails KTO by 2.50 to 15.92 points. CM-GSPO improves on KTO by 9.00 points on BrowseComp, 12.61 on Seal-0, and 8.00 on Xbench, but degrades BrowseComp-ZH accuracy from 75.78\% to 71.97\%, a decrease of 3.81 percentage points. Thus, the online-training gains are not uniform across benchmarks. The final BrowseComp-ZH system result with self-verification is 78.20\%; it should not be attributed to CM-GSPO training alone. Its margin over Compaction RL ranges from 11.50 to 19.82 points, and the largest margin is on Seal-0.
 
All sub-trajectories share the rollout-level advantage (Section~\ref{sec:cm_gspo}). Consistent with~\cite{pang2026glasseek}, these results support the combined CM-GSPO configuration over the compaction-only baseline. The table does not ablate sub-trajectory clipping and advantage reweighting individually, so it does not isolate their separate contributions. The scores are point estimates under one fixed protocol, so we treat them as directional evidence.

\FloatBarrier
\section{Limitations}
\label{sec:limitations}

\paragraph{Evaluation limitations.}
Deep search evaluation depends on changing webpages, search ranking, geographic availability, and tool failures. Even a fixed benchmark can therefore exhibit temporal variance. Public benchmark questions may also appear in pretraining data or online repositories, so leakage cannot be ruled out. LLM judges introduce additional uncertainty despite prompt freezing and human audits.

\paragraph{Verification and harness effects.}
Verification during testing is essential: checking final answers against cited evidence and inspecting trajectory-level tool use helps distinguish genuine search competence from accidental matches, stale pages, or judge errors. The surrounding agent harness can materially affect benchmark outcomes. Search and fetch tools, prompt templates, context-compression rules, stopping criteria, retry policies, tool budgets, and judge configuration all shape the trajectory and final score; changing these components can produce large differences even with the same checkpoint. We therefore treat reported numbers as measurements of the model--harness system under a specified protocol, and comparisons are meaningful only when these components are held fixed.

\FloatBarrier
\section{Conclusion}
\label{sec:conclusion}

We presented \agentname, a full-stack approach for training and evaluating 9B and 27B deep search agents. The method combines verifiable question synthesis, evidence-grounded trajectory curation, SFT, KTO-based offline preference optimization, and CM-GSPO for online learning under context compression. The data pipeline makes difficult questions auditable by requiring unique answers, explicit evidence chains, and verification of competing candidates, while the post-training stack turns those properties into search behavior that can be evaluated under a fixed harness.

Under our evaluation protocol with self-verification, \modelnine{} achieves 66.51\%, 67.82\%, 59.00\%, and 61.26\% on BrowseComp, BrowseComp-ZH, Xbench, and Seal-0, respectively. \modeltwentyseven{} reaches 80.49\%, 78.20\%, 75.00\%, and 65.77\% on the same benchmarks. The results show that \modelnine{} exceeds the available scores of the listed open-source baselines at or below 9B across the four benchmarks, while \modeltwentyseven{} remains competitive with substantially larger models. 

In conclusion, these results show that reliable deep-search performance depends on the interaction between verifiable data construction, evidence-sensitive preference optimization, and context-aware online learning, rather than on model scale alone. The improvements over the base models on all four benchmarks in the main results, at both model sizes, support the generality of this quality-first recipe. These observations align with our aim of developing \agentname{} for improving long-horizon searching under a controlled evaluation protocol.

\clearpage
\bibliographystyle{unsrtnat}
\bibliography{main}

\clearpage
\FloatBarrier
\appendix

\clearpage

\section{Training Details}
\label{app:training_details}

Tables~\ref{tab:sft_training_details}, \ref{tab:preference_training_details}, and~\ref{tab:cm_gspo_training_details} summarize the training configurations. TP, PP, CP, VPP, and SP denote tensor, pipeline, context, virtual pipeline, and sequence parallelism, respectively; the following numeral gives the configured degree. FSDP2 denotes Fully Sharded Data Parallel version 2. PPO and GRPO denote Proximal Policy Optimization and Group Relative Policy Optimization.

\begin{table}[H]
\centering
\small
\caption{Supervised fine-tuning launch configurations.}
\label{tab:sft_training_details}
\begin{tabular}{p{0.24\linewidth}p{0.33\linewidth}p{0.33\linewidth}}
\toprule
Setting & 9B & 27B \\
\midrule
Base checkpoint & Qwen3.5-9B & Qwen3.6-27B \\
Training framework & ms-swift + Megatron-LM & ms-swift + Megatron-LM \\
Fine-tuning mode & Full-parameter & Full-parameter \\
Numerical precision & bfloat16 & bfloat16 \\
Maximum training sequence length & 131,072 (128K) & 131,072 (128K) \\
Model maximum position embeddings & 262,144 & 262,144 \\
Micro / global batch size & 1 / 64 & 1 / 64 \\
Learning rate / schedule & $2\times10^{-5}$; min $2\times10^{-6}$; 3\% warmup & $1\times10^{-5}$; min $1\times10^{-6}$; 3\% warmup \\
Parallelism topology & TP2 / PP2 / CP4 / VPP2 & TP4 / PP2 / CP4 / VPP2 \\
Long-sequence optimization & Packing, padding-free, full recompute, Flash Attention & Packing, padding-free, full recompute, Flash Attention \\
\bottomrule
\end{tabular}
\end{table}
\FloatBarrier

\FloatBarrier

\begin{table}[H]
\centering
\small
\caption{Preference-optimization launch configuration.}
\label{tab:preference_training_details}
\begin{tabular}{p{0.24\linewidth}p{0.33\linewidth}p{0.33\linewidth}}
\toprule
Setting & Main route: KTO & Selection baseline: DPO \\
\midrule
Initialization &  SFT checkpoint & SFT checkpoint  \\
Source pool & Self-generated feedback rollouts + teacher-positive/model-negative contrast trajectories & Same source pool \\
Training view & Unpaired desirable/undesirable trajectories & Paired chosen/rejected trajectories \\
Numerical precision & bfloat16 & bfloat16 \\
Data scale & 10,672 trajectories; 5,336 desirable / 5,336 undesirable & 7,080 effective pairs \\
Maximum training sequence length & 131,072 (128K) & 131,072 (128K) \\
Learning rate / warmup & $5\times10^{-7}$ / 5\% & $5\times10^{-7}$ / 5\%\\
Micro / global batch size & 1 / 16 &  1 / 16 \\
Parallelism & 9B: TP2 / PP2 / CP4 / VPP2; 27B: TP4 / PP2 / CP4 / VPP2 & 9B: TP2 / PP2 / CP4 / VPP2; 27B:  TP4 / PP2 / CP4 / VPP2\\
$\beta$ & 0.1 & 0.1 \\
Long-sequence optimization & Packing; full recompute; Flash Attention & Packing; full recompute; Flash Attention \\
\bottomrule
\end{tabular}
\end{table}
\FloatBarrier

\begin{table}[H]
\centering
\small
\caption{CM-GSPO launch configurations.}
\label{tab:cm_gspo_training_details}
\begin{tabular}{p{0.24\linewidth}p{0.33\linewidth}p{0.33\linewidth}}
\toprule
Setting & 9B CM-GSPO & 27B CM-GSPO \\
\midrule
Initialization
    & KTO checkpoint
    & KTO checkpoint \\
  Learning rate / schedule
    & $1\times10^{-6}$ / constant
    & $1\times10^{-6}$ / constant \\
  Training batch / rollouts per prompt
    & 16 / 8
    & 16 / 8 \\
  PPO mini-batch / physical mini-batch
    & 16 / 128
    & 16 / 128 \\
  Actor / rollout parallelism
    & FSDP2 with Ulysses SP4 / tensor parallelism 4
    & FSDP2 with Ulysses SP4 / tensor parallelism 4 \\
  Context / prompt / response length
    & 131{,}072 / 4{,}096 / 8{,}192 tokens
    & 131{,}072 / 4{,}096 / 8{,}192 tokens \\
  Loss / advantage estimator
    & GSPO (sub-trajectory-level ratio), sequence-mean token-mean / GRPO
    & GSPO (sub-trajectory-level ratio), sequence-mean token-mean / GRPO \\
  Clipping / importance correction
    & $3\times10^{-4}$ and $4\times10^{-4}$ / token-level, threshold 2.0
    & $3\times10^{-4}$ and $4\times10^{-4}$ / token-level, threshold 2.0 \\
  Compression
    & Threshold 0.5; target ratio 0.2; protect first 3 and last 20 messages
    & Threshold 0.5; target ratio 0.2; protect first 3 and last 20 messages \\
    Sub-trajectory advantage reweighting
      & Enabled; $\widehat A^{\mathrm{cm}}_{i,k}=\tfrac{L_{i,k}}{L_i}\widehat A_i$ with rollout-level $\widehat A_i$
  shared across segments
      & Enabled; $\widehat A^{\mathrm{cm}}_{i,k}=\tfrac{L_{i,k}}{L_i}\widehat A_i$ with rollout-level $\widehat A_i$
  shared across segments \\
    Summarization tokens
      & Masked out of the RL loss (trainable trace types: agent only)
      & Masked out of the RL loss (trainable trace types: agent only) \\
  Validation / checkpointing
    & 32 examples; before training and every 25 steps / every 10 steps, keep 1
    & 32 examples; before training and every 25 steps / every 10 steps, keep 1 \\
\bottomrule
\end{tabular}
\end{table}

\section{BrowseComp Reference-Answer Review}
\label{app:browsecomp_ambiguity}

During judging, we found a number of errors in the reference answers of the \href{https://openaipublic.blob.core.windows.net/simple-evals/browse_comp_test_set.csv}{official BrowseComp test set}~\cite{wei2025browsecomp}: some reference answers are misspelled, and others are not the only valid answer to the question as written. This appendix lists these cases, distinguishing answer normalization from possible question ambiguity. We use the Agents-A1 LLM-based judging procedure and evaluation prompts released in its Search evaluation code\footnote{\url{https://github.com/InternScience/Agents-A1/tree/main/evaluation/Search}; judge prompts: \url{https://github.com/InternScience/Agents-A1/blob/main/evaluation/Search/judger/prompt.py}}. In our evaluation, all answer variants and the alternative candidate listed in Table~\ref{tab:browsecomp_golden_errors} were judged correct and counted as correct in the reported scores. Under these Agents-A1 grading rules, punctuation differences, equivalent name forms, and minor name typos that preserve identity do not by themselves change correctness. A distinct candidate answer, such as the one discussed for entry 1018, concerns question ambiguity rather than spelling normalization. Entry indices are zero-based.

\paragraph{An underspecified question.}
The reference answer for entry 1018 is \textbf{Rishi Kapoor}, but \textbf{Matthew Avery Modine} also fits the question.

\paragraph{Original question.}
\begin{quote}
\small
I am seeking the full name of an actor based on the following details:
\begin{itemize}
    \setlength{\itemsep}{2pt}
    \setlength{\parskip}{0pt}
    \item This actor's father was born between 1920 and 1930 (exclusive of the endpoints).
    \item This actor's father got married between 1940 and 1950 (exclusive of the endpoints).
    \item This actor was born between 1950 and 1960 (exclusive of the endpoints).
    \item The actor went to a school that was founded between 1940 and 1950 (exclusive of the endpoints).
    \item This actor married between 1975 and 1985 (exclusive of the endpoints) and as of 2023 has two children.
\end{itemize}
Can you specify the full name of the actor?
\end{quote}

The phrase ``went to a school'' leaves the institution's type unspecified, allowing a professional acting school to satisfy the clue. Modine attended the Stella Adler Studio of Acting, founded in 1949~\cite{nyu_stella_adler_alumni,stella_adler_history}, and our review of all five clues found that he satisfies them.

\paragraph{Reference-answer variants and ambiguity.}
Table~\ref{tab:browsecomp_golden_errors} summarizes all eight reviewed entries, including entry 1018 discussed above. For entry 1018, Rishi Raj Kapoor is a fuller name of the reference actor, whereas Matthew Avery Modine is a distinct candidate identified in our ambiguity review.

\begin{table}[H]
\centering
\small
\caption{BrowseComp reference-answer variants and an alternative candidate identified during judging. All listed variants and the alternative candidate were judged correct in our evaluation using the Agents-A1 judging procedure.}
\label{tab:browsecomp_golden_errors}
\begin{tabular}{@{}l>{\raggedright\arraybackslash}p{0.29\linewidth}>{\raggedright\arraybackslash}p{0.29\linewidth}>{\raggedright\arraybackslash}p{0.27\linewidth}@{}}
\toprule
Entry & Reference answer & Answer judged correct & Reason \\
\midrule
260 & Examination of Prospective Teachers' Creative Comparisons for the Concept of \textquotedbl{}Science Education'' & The same title with matching quotation marks & The two differ only in punctuation. \\
\addlinespace
324 & Arumah Oteh & Arunma Oteh & The reference answer is misspelled; Arunma Oteh is the individual's documented name. \\
\addlinespace
405 & Masata Kato & Masato Kato & The reference answer is misspelled. \\
\addlinespace
444 & multiple-membership, multiple-classification & multiple-membership, multiple-classification multilevel model & The added words describe the model in the paper and do not change the expansion. \\
\addlinespace
637 & Parle & Parle-G (Parle Products) & This names the product and its parent company, identifying the brand more specifically. \\
\addlinespace
851 & La Estrella; Argentina & Altos de la Estrella / La Estrella; Argentina & Altos de la Estrella is the fuller name of the same building. \\
\addlinespace
994 & Nagkahiusang Manunulat sa Cagayan de Oro & Nagkahiusang Magsusulat sa Cagayan de Oro & Sources identify both as the same organization (NAGMAC). \\
\addlinespace
1018 & Rishi Kapoor & Rishi Raj Kapoor; Matthew Avery Modine & The former is a fuller name of the reference actor; the latter is the distinct candidate discussed above. \\
\bottomrule
\end{tabular}
\end{table}
\end{document}